# Forecasting the Short-Term Energy Consumption Using Random Forests and Gradient Boosting

Cristina Bianca Pop, Viorica Rozina Chifu, Corina Cordea, Emil Stefan Chifu, Octav Barsan
*Computer Science Department*
*Technical University of Cluj-Napoca*
Romania
{viorica.chifu, cristina.pop}@cs.utcluj.ro

*Abstract*—This paper analyzes comparatively the performance of Random Forests and Gradient Boosting algorithms in the field of forecasting the energy consumption based on historical data. The two algorithms are applied in order to forecast the energy consumption individually, and then combined together by using a Weighted Average Ensemble Method. The comparison among the achieved experimental results proves that the Weighted Average Ensemble Method provides more accurate results than each of the two algorithms applied alone.

*Index Terms*—forecasting the energy consumption, random forests, gradient boosting, ensemble method

## I. INTRODUCTION

Electrical energy is one of the essential resources that ensures many basic human needs and maintains our society at the civilization and comfort levels we are used to. However, the continuous increase of demand and consequently the increase of production started to raise concerns about a series of detrimental environmental and economic effects, together with problems on the exhaustion of energy sources and supply. For example, in developed countries, residential and commercial buildings gained an increased influence over the total energy consumption, their contribution reaching values between 20% and 40%; thus they surpass other substantial sectors such as the industry and the transportation. Among the building services, the ones which shown the greatest increase lately are the heating, ventilation and air conditioning (HVAC) systems. In this context, several strategies have been proposed to increase the global energy efficiency, as follows: the energy retrofitting for homes and buildings (e.g. proper insulation); reducing the cooling loads in buildings through energy-efficient design/ passive cooling strategies/ usage of energy-efficient cooling equipments; raising the public awareness on the importance of efficient energy use etc. [16]. All these can be complemented by the ability of the electricity supplier companies to forecast the energy demand and plan its generation accordingly. The forecasting can also be profitable for the consumers, especially for the industrial ones, who buy day or hour ahead a certain amount of energy for a better price, but if not enough is bought or more energy is needed this will involve a financial loss.

This paper investigates two machine learning algorithms, namely Random Forests and Gradient Boosting, when they are used individually and then used as combined in a Weighted Average Ensemble Method for forecasting the energy consumption based on historical data. The obtained experimental results prove that the Weighted Average Ensemble Method provides more accurate results than each of the two algorithms applied alone.

The rest of the paper is structured as follows. Section II presents related work. Section III describes the approach we propose for forecasting the energy consumption using Random Forests and Gradient Boosting, while section IV discusses experimental results. We end our paper with conclusions.

## II. RELATED WORK

In the field of forecasting the energy consumption, there are many approaches available in the research literature that focus on the short-term prediction [1]- [4], [6], [8], [9] - [15]. In what follows, we present some of the most relevant approaches in this field.

In [15], Feed-Forward Back-Propagation neural networks (FFBP) and Decision Trees (DT) are used for short-term prediction of HVAC energy consumption in a hotel. The prediction was made by taking into account the outdoor temperature, the humidity, the speed of the wind, the hour of the day, the day of the week, the month of the year, the number of guests in a day, and the number of booked rooms. The results prove that the FFBPs provide predictions of higher accuracy than the DTs do. [1] proposes an approach for forecasting the electricity load for the day-ahead in micro-grids. It is based on Self-Recurrent Wavelet Neural Networks (SRWNN). The authors use as case study a smart power micro grid from the British Columbia Institute of Technology in Vancouver, which includes power plants, campus loads, as well as command, control and communication networks. To asses the performance of the SRWNNs, they compare the SRWNNs on one hand with the Wavelet neural networks and the Multi-Layer Perceptron neural networks on the other hand. Similarly, for short term forecasting of energy usage in commercial buildings, [3] uses artificial neural networks combined with the Bayesian regularization algorithm. As potential predictors, 9 independent variables are considered, which are classified in three main categories: environmental variables, time indicators, and operational conditions. Based on the experimental results, the authors conclude that the most important variables for predicting the energy usage on short term are: the day type, the time of the day, the temperature schedule of the HVAC set, the outdoor air temperature, and the humidity. The authors of [6] propose a hybrid approach that combines support vector machines (SVMs) with the Differential Evolution in order to predict the short-term and medium-term energy consumption in institutional buildings. The SVM is used in order to build the forecasting models, while the Differential Evolution helps in identifying the optimal configuration of weights for each model. The prediction was made for half an hour and for a whole day. For short term prediction of the electricity consumption of appliances in a domestic building, in [7] the Multiple Linear Regression, the SVMs, the Random Forests (RFs), and the Gradient Boosting Machines (GBMs) are analyzed. The most accurate predictions are obtained with the GBM and RF algorithms. Also, it was noticed that the weather data and the atmospheric pressure were the most important variables to influence the prediction accuracy. In [5], the adaptive neuro-fuzzy inference system (ANFIS) is used for predicting the energy consumption in residential buildings. The predicted energy consumption is estimated based on the following parameters: the structure of the building, the value of the insulation, and the thickness of the insulation. The accuracy of prediction when using ANFIS has been compared to the one obtained when using Artificial Neural Networks and Genetic programming. The experimental results prove that the ANFIS provides more accurate results than the Artificial Neural Networks and the genetic programming. In [4], two neural neworks are analyzed comparatively, when it comes to make short-term predictions of the energy consumption at district-level. The first one uses One-Step Secant backpropagation, whereas the second uses BFGS Quasi-Newton backpropagation. The prediction is made by taking into account the energy consumption of the previous seven days as well as the environmental data (e.g. temperature, humidity, wind speed etc.). The authors compare the two neural networks with the Levenberg-Marquardt Backpropagation algorithm in terms of the predictions accuracy. The results prove that the two algorithms outperform the Levenberg-Marquardt backpropagation algorithm.

As compared with the forecasting methods enumerated here, our approach underlines the importance of the seasonal factors, such as season, day, month, and hour of the day, when it comes to improve the accuracy of the predictions. The seasonal factors are very relevant for residential buildings, since a big difference in terms of energy consumed was noticed in different seasons, on different days of the week, and at different hours of the day. Also, our approach uses a Weighted Average Ensemble Method that combines two individual algorithms in order to improve the forecasting results.

### III. FORECASTING THE ENERGY CONSUMPTION BY USING RANDOM FORESTS AND GRADIENT BOOSTING

The main steps of our approach are presented in Figure 1.

**Data pre-processing.** The operations involved in data preprocessing are data normalization, data aggregation, data splitting, and replacing the null values.

*Data normalization* brings the values of all data to the same scale. The methods used in our approach for scaling the values are Min-max scaler and MaxAbs scaler. The normalization is applied on training data.

For *data aggregation* we have used the following approach: for each entry in the data set we compute an index by dividing the total amount of time (in minutes) that has passed in a day up until the data entry has been recorded, over the granularity established for the predictions. The granularity of predictions is defined as the amount of time (in minutes) within which all the energy recordings are grouped together.

For *replacing the null values* we have used the interpolation based on neighboring values. The null values replacing is applied on training data.

For *splitting the data set* into the training set and the test set, we have used two approaches: chronological (i.e. ordered), and seasonal data splitting. In the case of *ordered splitting*, the only problem is that when having a restricted amount of data, there is a probability that the model will train only on some of the seasons. Then this trained model will make predictions on data corresponding to seasons for which no relevant training took place. The *seasonal data splitting* consists in taking a certain percentage from each month individually for training and then making the predictions on the remaining days of the month. By splitting the data in this way, we avoid the above mentioned problem encountered in the chronological (ordered) data splitting. Moreover, with seasonal data splitting we also make sure that the data used for training is diverse enough.

**Feature extraction.** The data set based on which the predictions will be made is composed of data entries, each of them being associated to a timestamp. Starting from these timestamps of the data entries, we have extracted the following features, which play an important role in the prediction: year, month, week of the year, day of the year, day of the month, day of the week, hour in the day, half hour in the day, season, weekend, and energy consumption values. In our experiments, we tested with different combinations of features, and we measured the accuracy of the predictions made for each combination of features.

**Training and forecasting.** In our approach, 80% of the data are used for training, and 20% for testing. For evaluation we have used the following metrics: the root mean square error (RMSE), the mean absolute error (MAE), the mean absolute deviation (MAD), and the mean absolute percentage error (MAPE).

### IV. EXPERIMENTAL RESULTS

This section presents the experimental results we achieved when applying the Random Forests and Gradient Boosting algorithms as well as a Weighted Average Ensemble method. The Weighted Average Ensemble method combines the predictions from the two models obtained using Random Forests and Gradient Boosting algorithms, where the contribution of each model is weighted proportionally to its quality (i.e. the accuracy of the model).

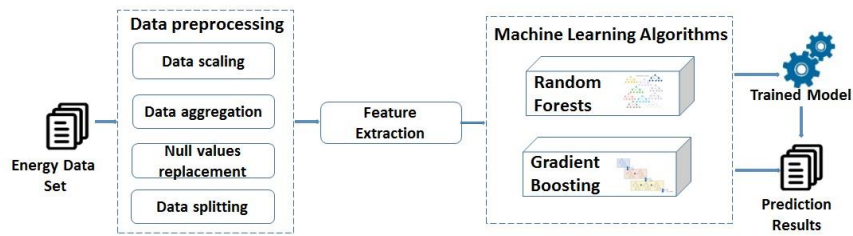

Fig. 1. The main steps of our approach

We used a data set in our experiments containing information about the energy power consumption of 6 HVAC chillers installed in a building, the power consumption being collected at every minute during one year [17].

Initially, we carried out a set of preliminary experiments in order to see how different granularities of predictions and different data splitting approaches influence the accuracy of the prediction.

A first aspect which has a big influence on the accuracy of the results is the granularity of the predictions, whose value is significant when aggregating the data within a specified amount of time, as well as when making the forecasting on the same amount of time. For data that describe a smaller amount of time, a smaller value for the granularity of predictions is also preferred, a good candidate being one hour. For data which cover a full year, it was noticed that a larger value for the prediction granularity, for example one day, is more suitable.

For data splitting, the first option was to allows the selection of the season to which the desired forecasting interval belongs. Hence, only data corresponding to that season will be used for training. It makes sense to have such an option, since training data from the summer, for instance, would not be relevant for predicting values in the winter.

The second data splitting method used in our experiments is to take a training set percentage from each season, and not from the whole year. Hence, data related to each period will be involved in the training process, which will increase the accuracy. Similarly, a third method chosen in the experiments takes a certain percentage from each month for training, which represents yet another way of taking samples from the whole interval available, in an evenly distributed way.

Figures 2-5 illustrate the results achieved when using Random Forests with the following configuration: 10 as maximum depth for trees, 15 trees (as number of trees), and 0.2 as minimum information gain threshold. The only difference among these graphs stems in splitting the data in different ways: ordered split (Figure 2), seasonal split (Figure 3), monthly split (Figure 4), and the spring season for training and prediction (Figure 5). The predictions are made with a granularity of one day, i.e. for an amount of time of one day.

Similarly, Figures 6-9 show the results achieved when using the Gradient Boosted Trees. The parameters we used for the Gradient Boosted Tree are 10 as maximum depth for trees and 0.2 as minimum information gain threshold. The predictions are made with a granularity of one day.

TABLE I
EXPERIMENTAL RESULTS ACHIEVED WITH RANDOM FORESTS, GRADIENT BOOSTING, AND WITH THE WEIGHTED AVERAGE ENSEMBLE METHOD

| Metric | Random Forest | Gradient Boosting | Weighted Average Ensemble Method |
|---|---|---|---|
| RMSE | 141.5 | 131.13 | 107.72 |
| MAE | 98.24 | 94.63 | 83.86 |
| MAD | 176.54 | 174.27 | 169.07 |
| MAPE | 17.06 | 15.89 | 14.27 |

We have also analyzed comparatively the experimental results obtained with Random Forests, Gradient Boosting, as well as with the Weighted Average Ensemble method. Table I illustrates the values of the evaluation metrics achieved with Random Forests, Gradient Boosting, and with the Weighted Average Ensemble method. Based on the values, we can conclude that the Weighted Average Ensemble method provides the best prediction results, i.e. the lowest values for the four above mentioned evaluation metrics.

## V. CONCLUSIONS

In this paper we presented a comparative analysis among two machine learning algorithms and a Weighted Average Ensemble method, as applied in the context of forecasting the energy consumption in public buildings. We have conducted a set of experiments, and we emphasized what happens with the results achieved when tuning certain configurations and parameters of the algorithms. Based on these experiments, we can conclude that the best results are obtained when using the Weighted Average Ensemble method, followed by Gradient Boosting. By carefully tuning the Gradient Boosting hyparameters, we obtain better results than with Random Forest, because in Gradient Boosting, unlike Random Forest (where the trees are independently builds and combined at the end), the trees are built sequentially and each new tree tries to correct the mistakes of the previous trees. However, the Gradient Boosting algorithm requires a longer training time than the Random Forest algorithm and is also more sensitive to over-adaptation phenomena in the case of noise data. In contrast with Random Forest and Gradient Boosting algorithms, Weighted Average Ensemble method combines the learned models with Random Forest and Gradient Boosting to get better results.

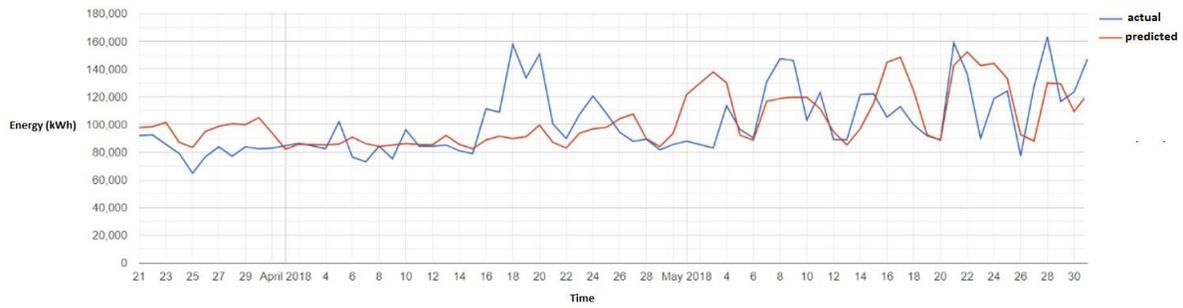

Fig. 2.  Predictions for a week using the algorithm for Random Forests - ordered data split

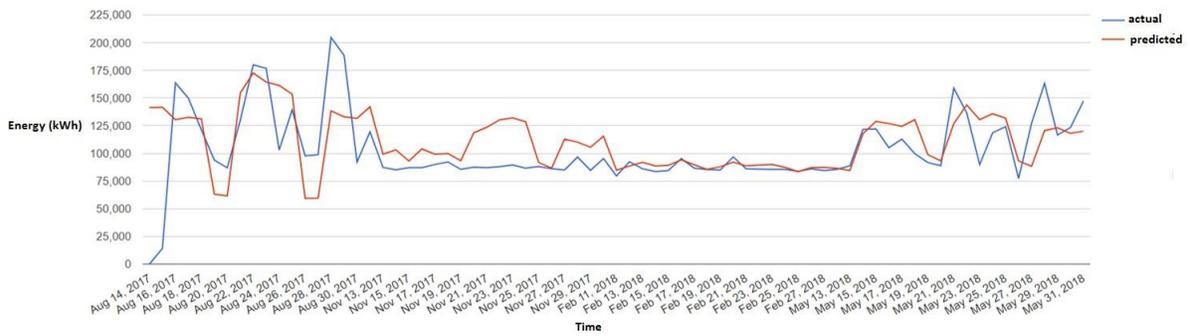

Fig. 3.  Predictions for a week using the algorithm for Random Forests - seasonal data split

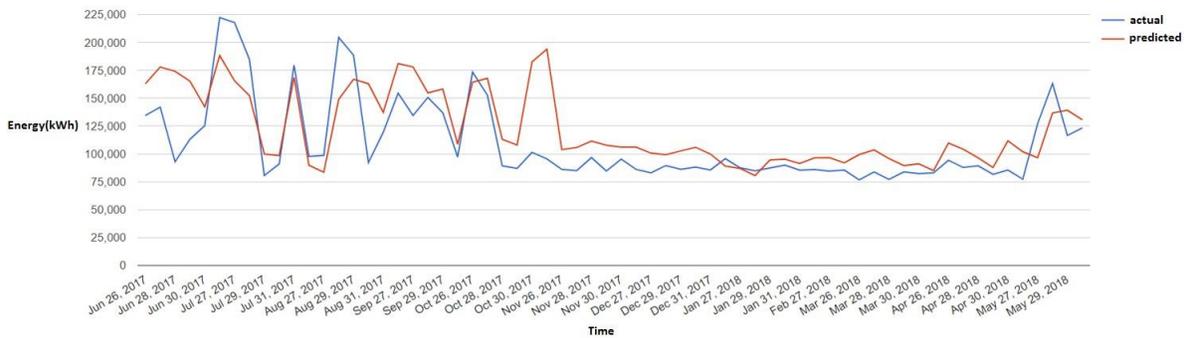

Fig. 4.  Predictions for a week using the algorithm for Random Forests - monthly data split

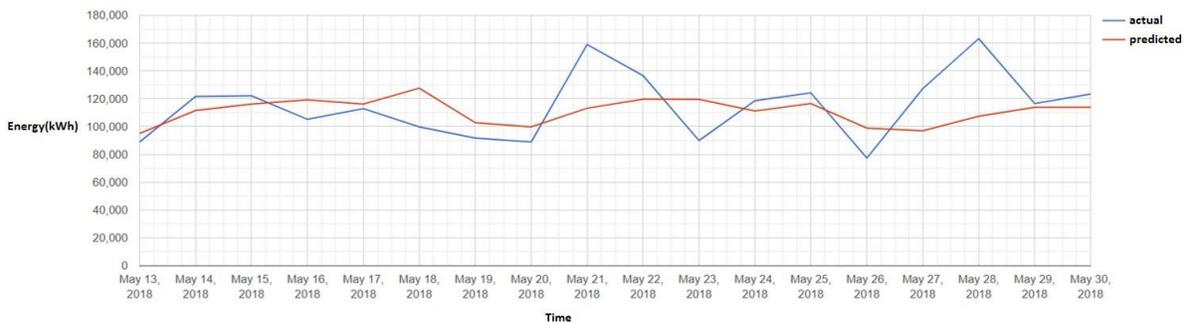

Fig. 5.  Predictions for a week using the algorithm for Random Forests - spring season data split

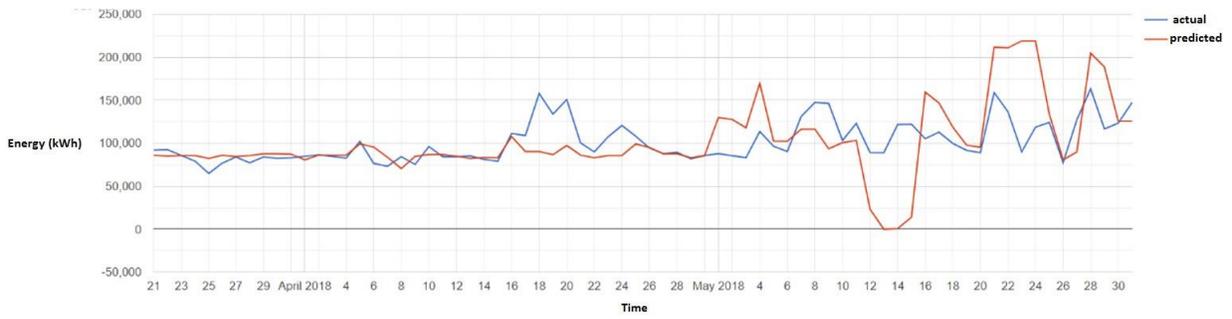

Fig. 6. Predictions for a week using the algorithm for Gradient Boosted Trees - ordered data split

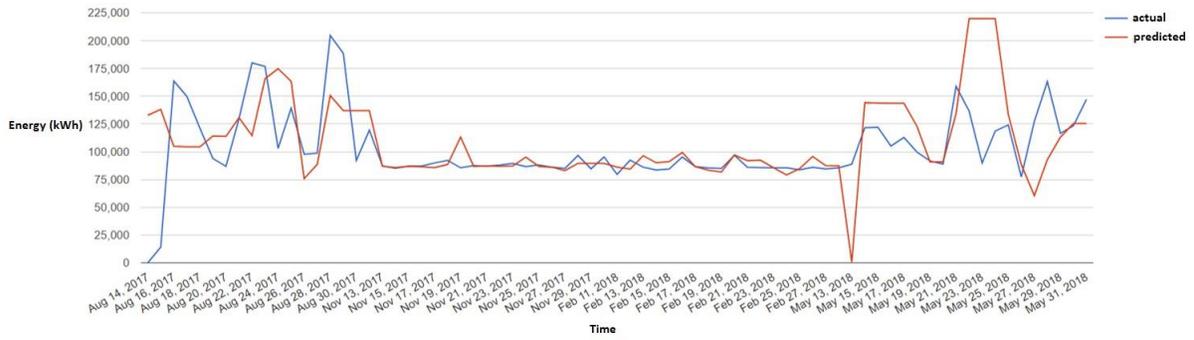

Fig. 7. Predictions for a week using the algorithm for Gradient Boosted Trees - seasonal data split

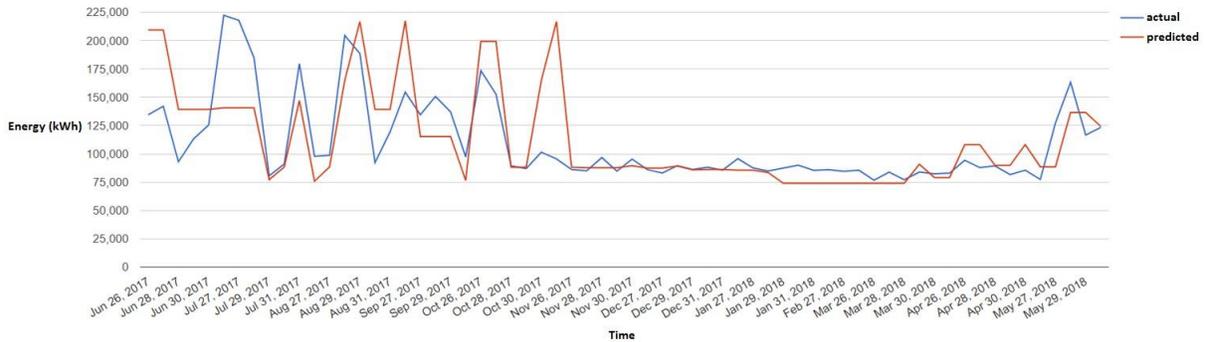

Fig. 8. Predictions for a week using the algorithm for Gradient Boosted Trees - monthly data split

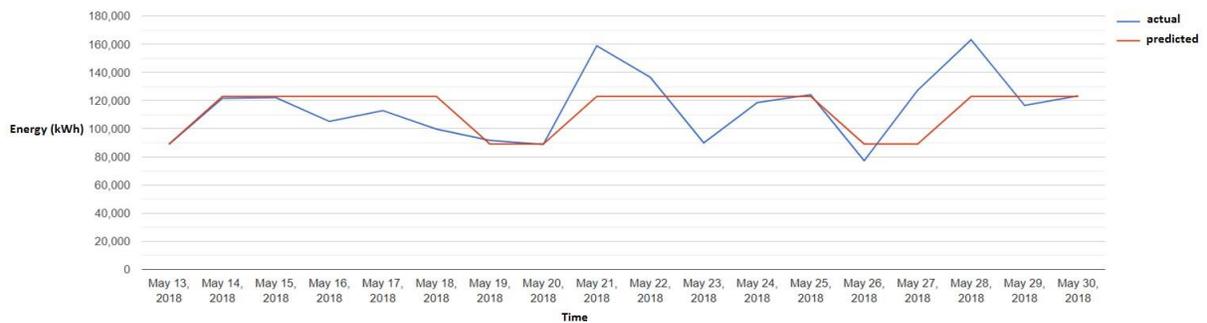

Fig. 9. Predictions for a week using the algorithm for Gradient Boosted Trees - spring season data split

ACKNOWLEDGMENT

This work has been conducted within the BRIGHT and eDREAM projects, grant number 957816 and 774478, funded by the European Commission as part of the H2020 Framework Programme.